\documentclass[letterpaper, 10 pt, conference]{ieeeconf}  

\IEEEoverridecommandlockouts                              

\usepackage{graphics} 
\usepackage{epsfig} 
\usepackage{mathptmx} 
\usepackage{times} 
\usepackage{amsmath} 
\usepackage{amssymb}  
\usepackage{booktabs}
\usepackage{multirow}
\usepackage{array}
\usepackage{graphicx}   
\usepackage{caption}
\usepackage{subcaption}

\title{\LARGE \bf
Dynamic Adaptive Legged Locomotion Policy via Decoupling Reaction Force Control and Gait Control}

\author{Renjie Wang$^1$, Shangke Lyu$^{2\dagger}$, Donglin Wang$^{1\dagger}$
\thanks{$^1$Renjie Wang and Donglin Wang are with Machine Intelligence Lab (MiLAB), School of Engineering, Westlake University, Hangzhou 310030, China.}
\thanks{$^2$Shangke Lyu is with Nanjing University, Suzhou, 215163, China.}
\thanks{$^{\dagger}$Corresponding author.}
}

\begin{document}

\maketitle
\thispagestyle{empty}
\pagestyle{empty}

\begin{abstract}
While Reinforcement Learning (RL) has achieved remarkable progress in legged locomotion control, it often suffers from performance degradation in out-of-distribution (OOD) conditions and discrepancies between the simulation and the real environments. Instead of mainly relying on domain randomization (DR) to best cover the real environments and thereby close the sim-to-real gap and enhance robustness, this work proposes an emerging decoupled framework that acquires fast online adaptation ability and mitigates the sim-to-real problems in unfamiliar environments by isolating stance-leg control and swing-leg control. Various simulation and real-world experiments demonstrate its effectiveness against horizontal force disturbances, uneven terrains, heavy and biased payloads, and sim-to-real gap. 
\end{abstract}

\section{Introduction}
\label{sec:introduction}





Achieving adaptive maneuver for legged robots closing to their real life counterparts in unstructured environments has been a long-term goal in the legged robot community. Reinforcement Learning (RL) has emerged as a cutting-edge control method for legged robot locomotion in recent years, and existing work has validated its flexibility in various real-world environments~\cite{hwangbo2019learning,lee2020learning,nahrendra2023dreamwaq,kim2025high}. However, the RL control paradigm suffers from significant performance degradation in out-of-distribution (OOD) conditions, as RL policies only have limited generalization ability in the regions near the state distributions that they have explored during training, and control performance becomes unreliable in unforeseen environments that differ from the pre-explored space. 


Legged locomotion typically involves two levels of control requirements. On the one hand, the controller must account for diverse contact forces to maintain balance during the stance phase. On the other hand, it must perform gait planning and adapt leg motions to abrupt state change (e.g., base velocity) and diverse terrains during the swing phase. However, in RL pipelines, these two objectives are tightly coupled: The action commands sent to the PD controller are expected to simultaneously generate appropriate reaction forces and leg movements. While this coupling simplifies controller specification, it substantially enlarges the state exploration space by forcing the policy to balance two distinct objectives, thereby increasing both the difficulty and cost of learning. More critically, in our view, this coupling is a key reason for the discrepancy between simulation and reality during policy deployment, and it makes online policy updates difficult. Specifically, the tight linkage between force control and position control means that changes in either reaction forces or leg movements can easily disrupt the delicate balance encoded in the learned action sequences, leading to a more severe sim-to-real gap. This also poses barriers to efficient online fine-tuning, since policy updates must simultaneously address both force regulation and motion generation.    

\begin{figure}[t]
  \centering
  \includegraphics[width=1\linewidth]{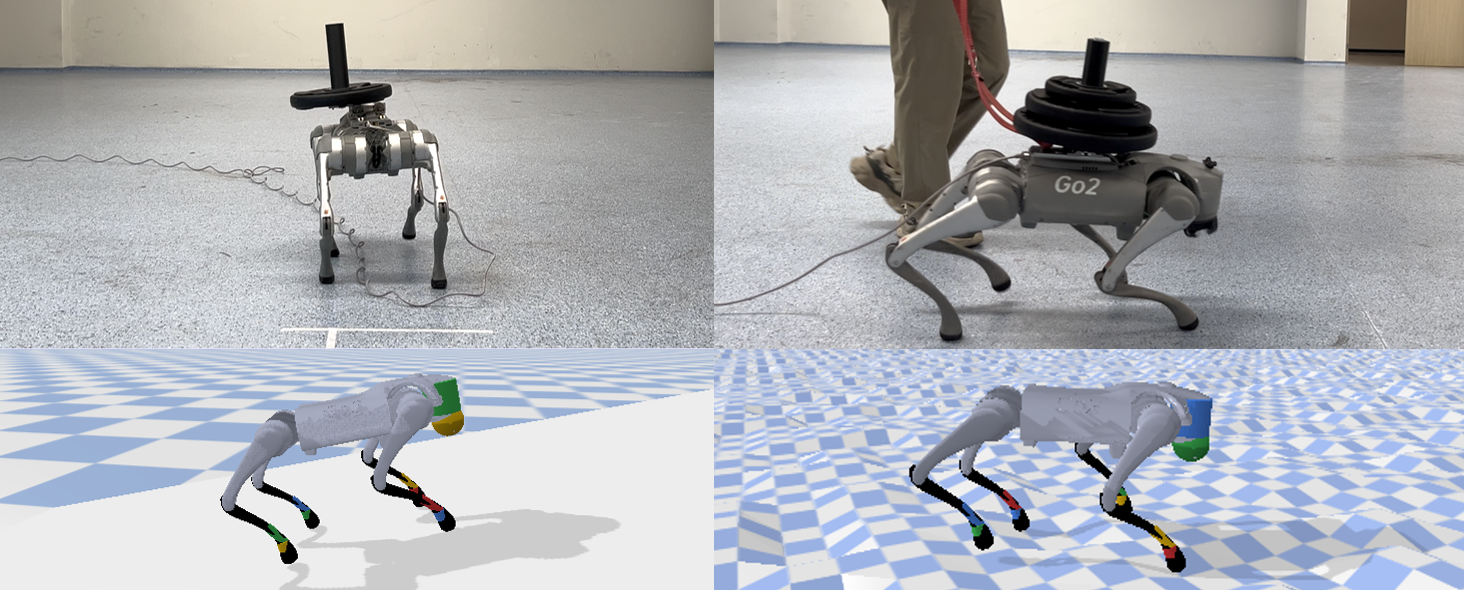} 
  \caption{Our proposed method can handle various unknown and unfamiliar environments and disturbances.}
  \label{fig:show_photo}
\end{figure}

Currently, a mainstream approach to alleviate OOD and sim-to-real issues are to introduce domain randomization (DR)~\cite{lee2020learning,tan2018sim,rudin2022learning} in both simulation environments and interaction dynamics, where the parameters are randomized during training. Therefore, this approach enriches the diversity of simulation environments and robot dynamics to best cover real environments, and it is effective because if we have a control policy that can handle an ensemble of different environments, the real-world environments are likely to be included~\cite{ha2025learning}. However, gaining the ability to handle a larger bundle of different environments sacrifices the optimality for the individual environment, which is commonly represented by a drop in reward profiles~\cite{xiao2025learning}.

Regardless of DR, a feasible way to mitigate the sim-to-real issue is to partially decouple force control and position control, thereby isolating their interdependence when dynamics or environmental conditions change. This decoupling also enables efficient online fine-tuning. For instance, the reaction force can be adjusted to account for dynamics variations without interfering with the action commands for leg motion, which may significantly enhance the robustness and adaptability of the learned policy. Inspired by~\cite{lyu2024rl2ac}, where a well-trained RL policy is interpreted as a combination of feedforward and feedback components similar to the MPC framework for contact-rich legged locomotion, we adopt a different perspective. Instead of relying on a single RL policy to infer both feedforward and feedback inputs, we propose a novel framework that decouples feedforward force control and feedback motion control, assigning them to MPC and RL respectively, thus enabling both efficient learning and dynamic adaptation.

More specifically, we alternate the PD controller used in the traditional RL-based control framework with a low-level controller (LLC) that integrates feedforward control and feedback control that resembles the model-based control scheme~\cite{kim2019highly,di2018dynamic}. The feedforward dynamics is provided by MPC and the RL agent mainly focuses on inferring appropriate target joint positions tracked by the PD feedback control. Finally, the joint torque commands sent to motors are composed of the feedforward torque and the feedback torque generated by PD controller. The MPC module is running at a high frequency to adapt to unknown disturbances and the RL policy is used to track the gait. It is also worth noting that MPC does not need to provide highly precise feedforward torques; instead, a coarse baseline is often sufficient, as the policy can further improve upon it through self-learning. This reduces the reliance of MPC on an accurate dynamics model and accelerates online optimization. 

In summary, the contributions of this work are as follows:
\begin{itemize}
\item We propose a hybrid scheme that decouples reaction force control from the RL policy. This allows the RL policy to focus primarily on motion tracking, leading to more efficient policy learning and a reduced sim-to-real gap.
\item To enhance robustness, we complement RL with an online MPC module that enables real-time adaptation to unforeseen environments and disturbances during deployment.
\item Extensive experiments demonstrate that our method achieves robust and reliable task execution in unfamiliar environments and out-of-distribution (OOD) task scenarios.
\end{itemize}

\section{Related Work}


\subsection{RL-based Robust Legged Locomotion}
In order to achieve zero-shot deployment of RL policies on legged robots, domain randomization (DR) has been widely used in many studies~\cite{margolis2023walk,han2024lifelike,kumar2021rma}, which randomizes environment dynamics, robot commands, external disturbances, etc., to maximum cover the real-world environments. In addition to DR, much of the work has used curriculum learning to master practical and dynamic skills step by step. \cite{lee2020learning} designed a delicate curriculum framework to choose a reasonable terrain difficulty according to the achievement of the given commands in the last episode; \cite{margolis2022rapid} leveraged a carefully synthesized curriculum framework for velocity command that decreases the range of lateral and turning velocity when heading velocity is high, which enables high-speed locomotion for MIT Mini Cheetah. 

In addition, training residual policies has been a kind of popular approach to perceive and take actions to compensate for nominal actions to resist unknown disturbances. \cite{zhang2025disturbance} proposed to use the additional disturbance observer and joint torque compensation networks to estimate external force and correct nominal actions, achieving to carry load surpassing the robots' weight. \cite{kim2025modular} proposed to use modular residual policies to compensate for the model mismatch of the footstep planner and the simplified dynamics of a MPC framework, respectively. Similarly, \cite{chen2024learning} leveraged a residual policy to compensate for the simplified dynamics and correct the tracking of the swing feet. We claim a difference of our decoupled framework from a residual policy approach, as the nature of a residual policy is 'compensation', while our method leverages MPC and RL modules to achieve feedforward control and feedback control components, respectively, through the insight of the characteristic and strength of different paradigms.

\subsection{Dynamic Adaptive MPC}
Due to the redundant design of quadruped robots in degree of freedom (DoF) that complicates the control problem, heuristic methods have been conventionally introduced for locomotion control, where the original system is represented by a simplified model, usually single rigid body dynamics (SRBD)~\cite{di2018dynamic,kim2019highly} or inverted pendulum model (IPM)~\cite{kang2022animal,chen2019optimal}. This simplification is practical due to the observation that the mass distribution of quadruped robots is concentrated at the base. The simplification of the system model significantly benefits MPC which solves receding-horizon optimization problems continuously, and enables MPC to react rapidly to unknown disturbances and finds the optimal actions to back to the desired trajectories. This online adaption is agnostic to surrounding environments and is only related to the design of the MPC formulation and its solving speed, which is complementary to RL-based approaches.

In addition, the robustness of MPC can be enhanced by solving robust optimization problems. One of the popular methods is to parameterize uncertainties as additional decision variables in the MPC formulation, for which robustness is achieved by solving a Min-Max problem that finds the optimal solution at the worst-case scenarios~\cite{zagorowska2024automatic,pandala2022robust,xu2023robust,trivedi2025chance}. Robustness can also be achieved using a tube-based MPC~\cite{chen2023quadruped}, which characterizes the capturability and constraints the final state within the tube.

\subsection{Decoupled Control for Legged Robots}
The idea of decoupling control for contact-rich quadrupedal locomotion tasks is widely adopted in robotics research. In traditional control methods, this idea is represented as the combination of feedforward force control and feedback motion control\cite{siciliano2009robotics}. Furthermore, MPC leverages this scheme by modeling legged robots as a switched system, which finds optimal feedforward components for stance leg by solving optimization problems and generates swing-leg kinematics trajectories for feedback components using Raibert's heuristic~\cite{raibert1986legged}.

For RL-based paradigms, it was wondered if the RL control policies follow similar rules. \cite{li2022bridging} observed that the RL control policies for a bipedal robot Cassie can be characterized as a linear model, which is decoupled between each dimension of the input-output imagination. \cite{lyu2024rl2ac} revealed that the PD controller used in RL-based approaches can be reformulated as a feedforward term and a feedback term, and this finding is then leveraged to design an adaptive control framework that compensates for the feedforward component to enhance the robustness of locomotion.  

\subsection{Combing Model-based Control and Model-free RL}
Much of the previous studies demonstrate the advantage of integrating model-based MPC and trajectory optimization (TO) with RL. Reference motion priors are generated using optimization-based methods to guide the learning of a locomotion policy in~\cite{jenelten2024dtc, kang2023rl+, wang2025integrating}, enabling efficient learning and distinct gait patterns. Some other work uses the learned RL policies to rectify model-based control~\cite{kim2025modular,chen2024learning,gangapurwala2022rloc}, and vice versa~\cite{lyu2023composite}.

\section{Preliminaries}

\subsection{RL for legged locomotion}
In this work, the RL environment is modeled as a Markov Decision Process (MDP), defined by the tuple \(M = (\mathcal{S}, \mathcal{A}, d_0, p, r, \gamma)\), where the state \(s \in \mathcal{S}\) and action \(a \in \mathcal{A}\) are continuous. The initial states of the environment follow a distribution \(d_0(s_0)\); the states progress with a state transition probability \(p(s_{t+1}|s_t,a_t)\); and each interaction is evaluated by a reward function \(r:\mathcal{S} \times \mathcal{A} \to \mathcal{R}\). \(\gamma\) is a discount factor defined in the range \(\gamma \in [0,1)\). 

A low-level PD controller is used in joint space by nominal RL-based approaches for legged locomotion, expressed as 
\begin{equation}\label{eq:nominal PD controller}
\boldsymbol{\tau} = \mathbf{K}_p(\mathbf{a}-\mathbf{q}) - \mathbf{K}_d \dot{\mathbf{q}},
\end{equation}
where the joint torque \(\boldsymbol{\tau} \in \mathbb{R}^{12}\) computed by the PD controller is sent to each motor; \(\mathbf{K}_p \in R^{12 \times12}\) and \(\mathbf{K}_d \in R^{12 \times12}\) are diagonal gain matrices; the action \(a \in \mathbb{R}^{12}\) represents the target joint positions, and \(\mathbf{q} \in \mathbb{R}^{12}\) and \(\dot{\mathbf{q}} \in \mathbb{R}^{12}\) denote the joint positions and velocities, respectively.

\subsection{Dynamics}
Let \(\mathbf{q}\) denotes the generalized coordinates, and the whole-body equation of motion (EoM) can be formulated as:
\begin{equation}\label{eq:EoM}
\mathbf{M}(\mathbf{q}) \ddot{\mathbf{q}} + \mathbf{C}(\mathbf{q}, \dot{\mathbf{q}})
= \mathbf{S}^T \boldsymbol{\tau}_{\text{motors}}
+ \mathbf{J}(\mathbf{q})^T \mathbf{f}_{\text{contact}}
\end{equation}
where on the left-hand side \(\mathbf{M}\) is the mass matrix and \(\mathbf{C}\) is the nonlinear term. The right-hand side contains the selection matrix \(\mathbf{S}\), motor actuation torques \(\boldsymbol{\tau}_{\text{motors}}\), contact Jacobian \(\mathbf{J}\), and contact forces \(\boldsymbol{f}_{\text{contact}}\).

\begin{figure*}[t]
  \centering
  \includegraphics[width=1\textwidth]{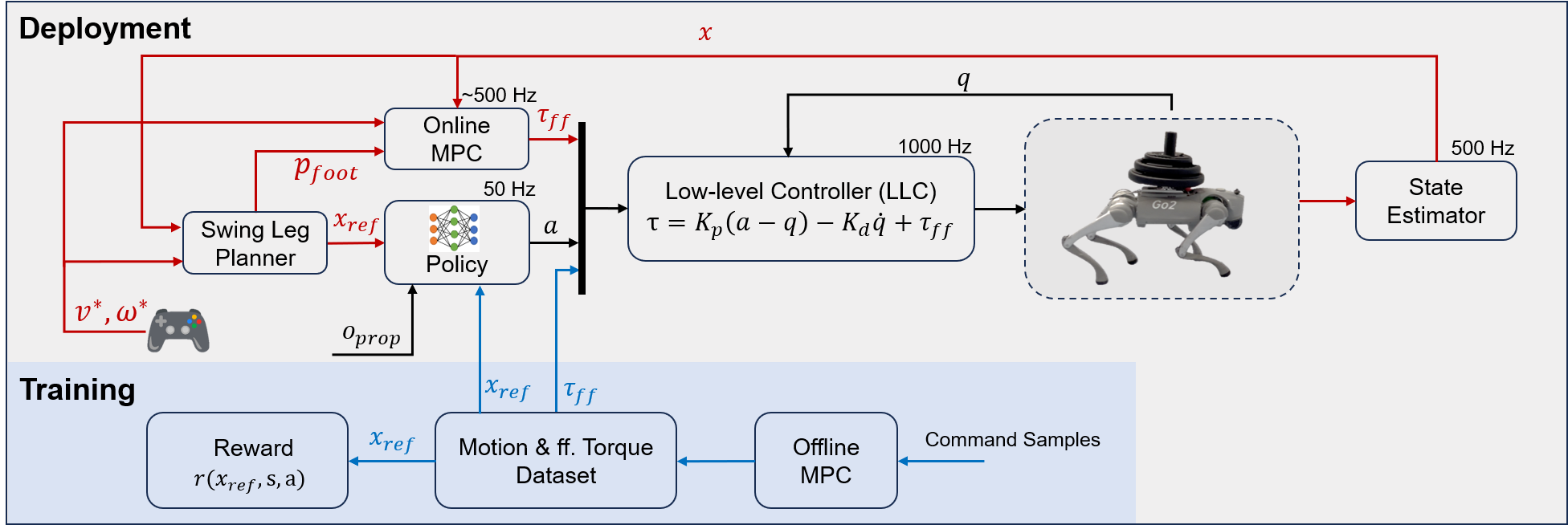} 
  \caption{Overview of the proposed method. The red line and blue line represent the data flow in deployment and in training, respectively. }
  \label{fig:framework}
\end{figure*}

\section{Method}

\subsection{System Overview}
This section provides a detailed description of the construction and training of our proposed method. As illustrated in Fig.~\ref{fig:framework}, the framework contains MPC and RL modules. Before training, the MPC module generates offline dynamics and kinematics trajectories, including joint torques, positions and velocities in generalized coordinate, and contact sequence and timing. Joint torque trajectories are directly fed into the low-level controller (LLC), and the reference motion is used to guide agent learning. During deployment, the MPC runs online to provide joint feedforward torques and motion reference.

\subsection{Robust MPC}
In this work a robust MPC controller from~\cite{trivedi2025chance}, namely Chance-Constraint MPC (CCMPC), is used to optimize ground reaction forces (GRF) and provide reference motion to the RL policies. CCMPC is a specific form of Stochastic Model Predictive Control (SMPC)~\cite{mesbah2016stochastic}, which directly incorporates uncertainties into controller design by modeling them as probability distributions and allows for a small probability of constraint violation. In addition, CCMPC is formulated as a quadratic programming problem (QP), enabling a fast control frequency. We emphasize that our method is not restricted to CCMPC; any MPC that finds optimized GRFs can be used in our framework.

Normally, MPC finds the optimal foot contact forces \(\mathbf{f}_{\text{contact}}\) in Equation~\eqref{eq:EoM} by solving the constrained optimal control problems. However, the modeling mismatch and the environmental information that cannot be encoded into the MPC formulation make the solution suboptimal. These uncertainties can arise from various factors, such as diverse payloads and terrains. CCPMC addresses the uncertainties using a stochastic variation of the SRBD model, and expresses the system as a discrete linear state-space equation:
\begin{equation}\label{eq:SRBD}
\begin{aligned}
\mathbf{x}_{i+1} &= \mathbf{A}_i \mathbf{x}_i + \mathbf{B}_i(\delta_i)\mathbf{u}_i + \mathbf{w}_i, \\
\text{where } \delta_i &\sim \mathcal{N}(\mathbb{E}[\delta_i], \Sigma_\delta), \quad 
\mathbf{w}_i \sim \mathcal{N}(\mathbb{E}[\mathbf{w}], \Sigma_w)
\end{aligned}
\end{equation}
where \(i\) denotes the discrete-time instance, \(\mathbf{A}_i\) is the state transition matrix, and \(\mathbf{B}_i\) is the imagination matrix that maps control inputs to state. \(\delta_i\) and \(\mathbf{w}_i\) represent the system and compensatory parametric uncertainties, respectively, that follow the Gaussian distribution. \(\mathbb{E}[\delta_i]=[m, \text{diag}(\mathbf{I}), \mathbf{r}_{i,1}^T,\ldots,\mathbf{r}_{i,4}^T]\) and \(\mathbb{E}[\mathbf{w}]=\mathbf{0}\) denote the mean of the corresponding distributions, respectively, where \(m\) is the robot mass, \(\text{diag}(\mathbf{I})\) represents the diagonal entries of the nominal inertia matrix, and \(\mathbf{r}_{i,k}\) for \(k \in [1,2,3,4]\) denotes the foot location. \(\Sigma_\delta\) and \(\Sigma_w\) are custom covariance matrix required to be tuned by users.

As mentioned earlier, CCMPC allows constraints violation within a small range to make distribution-based MPC feasible. More specifically, they are the friction cone constraints and the unilateral force constraints applied on GRF (i.e. \(\mathbf{u}_i\)), which are expressed as
\begin{equation}\label{eq:friction cone constraint}
\text{Pr} (\mathbf{C}_i\mathbf{u}_i \leq \mathbf{0}) \geq \epsilon,
\end{equation}
where \(\text{Pr}()\) stands for the probability that the event in the bracket will occur, \(\mathbf{C}_i \in \mathbb{R}^{20 \times 12}\) contains four friction cone constraints and one unilateral force constraint for each foot, resulting in 20 constraints in total, and \(\epsilon\) denotes acceptable probability thresholds for constraint satisfaction. There is an additional constraint in CCMPC to force the GRFs of the swing legs to be zero, which is expressed as
\begin{equation}\label{eq:swing GRF constraint}
\mathbf{D}_i \mathbf{u}_i = 0,
\end{equation}
where \(\mathbf{D}_i\) denotes the selection matrix for swing legs according to the desired contact state at the time instance \(i\).

Finally, the CCMPC can be formulated as follows:
\begin{equation}\label{eq:CCMPC formulation}
\begin{aligned}
\min_{\mathbf{x}_i, \mathbf{u}_i} \quad \mathbb{E} & \left[ \sum_{i=0}^{N-1} \| \mathbf{x}_{i+1} - \mathbf{x}_{\text{ref}, i+1} \|_{\mathbf{Q}}^2 + \| \mathbf{u}_i \|_{\mathbf{R}}^2 \right] \\
\text{s.t.} \quad & \text{Equation}~\eqref{eq:SRBD} \\
& \text{Equation}~\eqref{eq:friction cone constraint} \\
& \text{Equation}~\eqref{eq:swing GRF constraint}
\end{aligned}
\end{equation}
In Equation~\eqref{eq:CCMPC formulation}, the expectation is taken with respect to the distributions of the robot state and control inputs, namely \(\mathbf{x} \sim (\overline{\mathbf{x}}, \Sigma_x)\) and \(\mathbf{u} \sim (\mathbf{v}, \Sigma_u)\), where \(\overline{\mathbf{x}}\) and \(\mathbf{v}\) stand for the means, and \(\Sigma_x\) and \(\Sigma_u\) stand for the covariances. The CCMPC is conducted over the prediction N, with the semi-definite weight matrix \(\mathbf{Q}\) penalizing the tracking error between the state \(\mathbf{x}_i\) and the reference state \(\mathbf{x}_{\text{ref},i}\), and the positive definite weight matrix R penalizing the magnitude of the control input.

As multi-dimensional Gaussian probability density functions are integrated in Equation~\eqref{eq:CCMPC formulation}, it becomes nonlinear and computing-expensive. We only give a brief overview of the CCMPC formulation, for a detailed conversion of Equation~\ref{eq:CCMPC formulation} to a convex QP problem, refer to the original work~\cite{trivedi2025chance}.

\subsection{Motion Tracking RL Policy}
In our decoupled framework, we replace the nominal PD controller in Equation~\eqref{eq:nominal PD controller} with the following low-level controller (LLC):
\begin{equation}\label{eq:low-level controller}
\boldsymbol{\tau} = \mathbf{K}_{\text{p}}(\mathbf{\mathbf{a}-q}) - \mathbf{K}_{\text{d}} \dot{\mathbf{q}} + \boldsymbol{\tau}_{\text{ff}},
\end{equation}
where \(\mathbf{K}_{\text{p}}\) and \(\mathbf{K}_{\text{d}}\) represent the gain matrices and \(\mathbf{a}\) is the policy action. \(\mathbf{q}\) and \(\mathbf{\dot{q}}\) denote the feedback joint position and velocity, respectively. The feedforward joint torque \(\boldsymbol{\tau}_{\text{ff}}\) is computed by \(\boldsymbol{\tau}_{\text{ff}} = -\mathbf{J}^T\mathbf{f}_{\text{GRF}}\), where \(\mathbf{J}\) is the contact Jacobian matrix and \(\mathbf{f}_{\text{GRF}}\) is the GRF rolled out from MPC.

The architecture of the RL part follows our previous work~\cite{wang2025integrating}. The policy uses proprioceptive information, base velocity command, reference joint positions, reference leg contact state, and leg-phase left time as inputs, and outputs target joint positions used in the feedback control component. The observation space \(\mathbf{o}_t \in \mathbb{R}^{65}\) contains body angular velocity \(\boldsymbol{\omega}_t \in \mathbb{R}^3\), projected gravity \(\mathbf{g}_t \in \mathbb{R}^3\), base velocity command \(\mathbf{c}_t \in \mathbb{R}^3\), joint positions \(\mathbf{q}_t \in \mathbb{R}^{12}\), joint velocities \(\dot{\mathbf{q}}_t \in \mathbb{R}^{12}\), previous actions \(\mathbf{a}_{t-1} \in \mathbb{R}^{12}\), reference joint positions \(\mathbf{q}_{\text{ref},t} \in \mathbb{R}^{12}\), reference contact states \(\mathbf{s}_{\text{ref},t} \in \mathbb{R}^{4}\), and phase left time \(\mathbf{t}_{\text{phase},t} \in \mathbb{R}^{4}\). There is no additional privileged information used in the critic network. 
The reward is designed as the summation of tracking rewards and regularization rewards. The motion tracking terms contain the tracking of the reference base twist, linear and angular velocities, and joint positions, which can be formulated as   
\begin{equation}\label{eq:tracking rewards}
r_{x} = \exp \left( - \left\| \frac{x^*(t) \ominus x(t)}{\sigma_{x}} \right\|^2 \right),    
\end{equation}
where \(x(t)\) and \(x^*(t)\) are the measured and reference base pose, base velocity, or joint position, respectively; \( \ominus \) denotes the quaternion difference for the base orientation and vector difference otherwise. All of these motion priors are pre-generated by the MPC module for the corresponding velocity commands.
The regularization terms regulate the rate of change in actions and joint accelerations. The details of the reward functions are shown in Table~\ref{tab:reward_functions}.

\begin{table}[h]
\centering
\caption{Reward Functions}
\label{tab:reward_functions}
\scalebox{1.05}{
\begin{tabular}{lcl}
\toprule
Reward & Weight & Parameter \\ \midrule
Base rotation tracking (roll and pitch) & 0.5 & $\sigma=0.02$ \\
Base lin. velocity tracking & 1 & $\sigma=0.25$ \\
Base ang. velocity tracking & 0.5 & $\sigma=0.25$ \\
Joint position tracking & 1 & $\sigma=0.5$ \\
Action rate & $-5e^{-3}$ &  \\
Joint acceleration & $-1.25e^{-7}$ &  \\  \bottomrule
\end{tabular}}
\end{table}

\subsection{Implementation Details}
An \textit{Unitree Go2} quadruped robot is used in both training and deployment. Before training, a reference motion data set containing feedforward joint torques and kinematics trajectories is generated using CCMPC. The command velocities are uniformly sampled from the range \(v_{\text{x}} \in [-0.5, 1]\), \(v_{\text{y}} \in [-0.5, 0.5]\), \(\omega_{\text{z}} \in [-1, 1]\) with a step of 0.1, where \(v_{\text{x}}\), \(v_{\text{y}}\) and \(\omega_{\text{z}}\) denote the heading velocity, lateral velocity and turning velocity, respectively. The code and parameters of CCMPC are used the same as in the original work~\cite{trivedi2025chance} for both generating offline data set and online deployment. The prediction horizon and the planning time step are set as \(N=10\) and \(dt=0.02\), respectively, to align with the timestep of the RL policy. For each sample of the command velocity, trajectories are generated for one gait cycle, with the initial state set exactly the same as the desired state. 

Policies are trained on a plane, and during training, command velocities are randomly chosen from the pre-generated data set, and the corresponding feedforward joint torques and motion priors are used in the LLC and to guide the learning, respectively. The policies are trained in NVIDIA's Isaac Gym~\cite{makoviychuk2021isaac} and using the massively parallel RL framework~\cite{rudin2022learning} and PPO~\cite{schulman2017proximal}, with a NVIDIA A800 GPU. The environment setups (e.g., episode length and number of environments), training hyperparameters, and network parameters exactly follow~\cite{rudin2022learning} and train policies for 5,000 iterations. 

During deployment, there are three independent threads running at the same time for CCMPC, RL and main control loop, respectively. CCMPC runs online at approximately 500 Hz, to generate feedforward GRFs and reference motion, and the RL policy runs at 50 Hz. A Kalman-filter-based state estimator~\cite{katz2019mini} is implemented in the main control thread at 500 Hz. A laptop equipped with an NVIDIA GeForce RTX 3060 GPU and Intel Core i7-12700H CPU is used for the deployment.

\section{Results}

To evaluate the performance of our method, we train two more policies as baselines and for ablation studies, and all three methods use \(\mathbf{K}_{\text{p}}=28\mathbf{I}\) and \(\mathbf{K}_{\text{d}}=\mathbf{I}\) for all experiments, where \(\mathbf{I}\) denotes the identity matrix. The first policy, trained using the nominal RL framework with the PD controller in Equation~\eqref{eq:nominal PD controller}; the second policy is trained using the same method as the first one, with additional DR which is detailed in Table~\ref{tab:domain randomization}. To sum up, the policies used in the experiments are listed as follows:
\begin{itemize}
\item \textbf{Baseline}: Trained without feedforward control.
\item \textbf{Baseline-DR}: Trained without feedforward control and with additional DR.
\item \textbf{Ours}: Trained using the proposed method.
\end{itemize} 

\begin{table}[h]
\centering
\caption{Domain Randomization}
\label{tab:domain randomization}
\scalebox{1.05}{
\begin{tabular}{lccc}
\toprule
Methods & Friction & Mass (kg) & Max push vel. (m/s) \\ \midrule
Ours & [0.5, 1,25] & / & 1 \\
Baseline & [0.5, 1.25] & / & 1 \\
Baseline-DR & [0.5, 1.25] & \textbf{[-1, 3]} & \textbf{1.5} \\ \bottomrule
\end{tabular}}
\end{table}

\subsection{Simulation Experiments}
In the first part, we compare the performance of the three methods mainly in Pybullet~\cite{coumans2016pybullet} simulation environment.
\subsubsection{Effectiveness of the Decoupled Framework}
In this experiment, we command the robot to walk at 0.5 m/s on a plane ground using the three methods, respectively. 

\begin{figure}[t]
  \centering
  \includegraphics[width=1\linewidth]{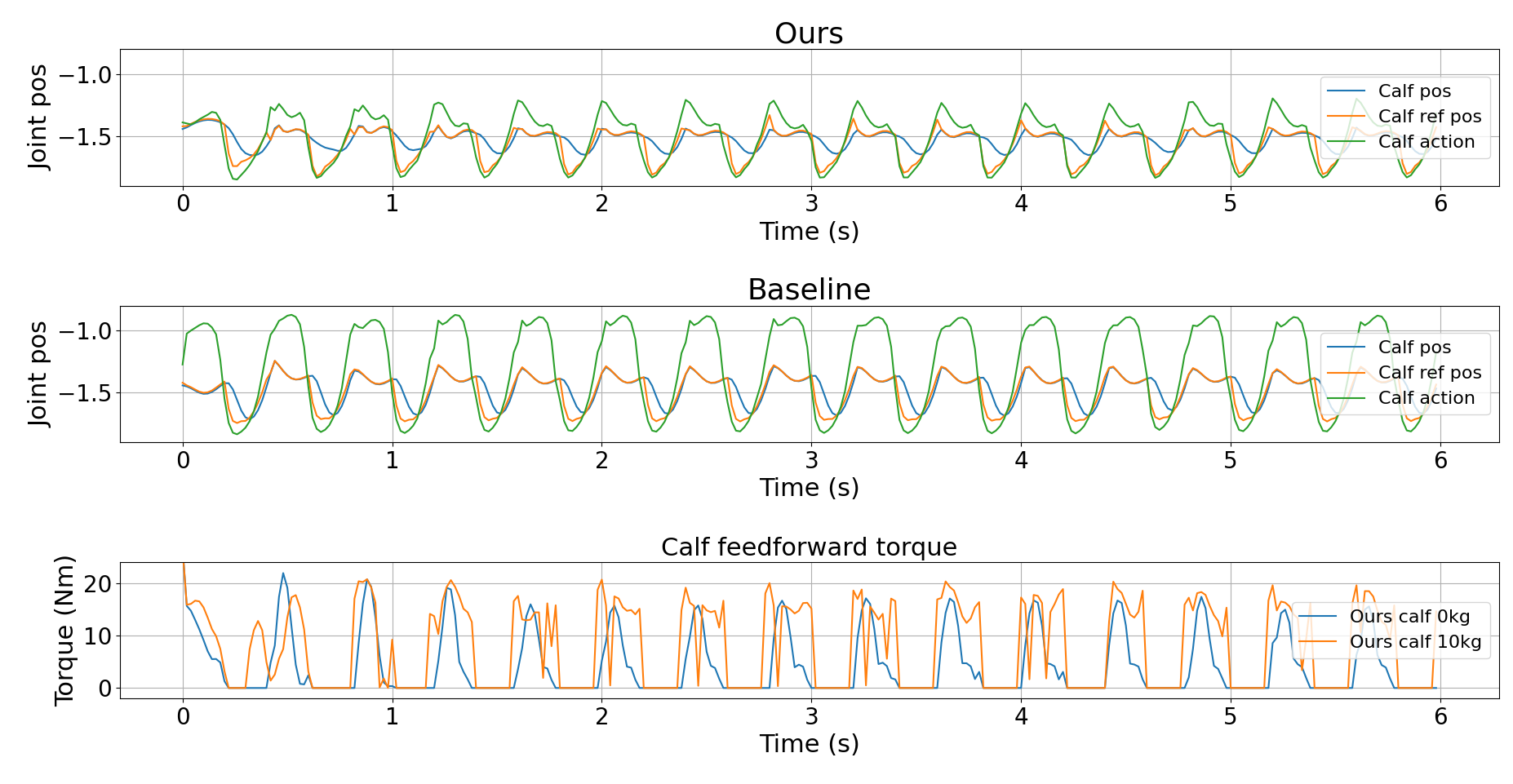} 
  \caption{Decoupling of dynamics and kinematics in FR calf joint space. The first two rows demonstrate the joint position, joint action, and reference joint position with our method and baseline, respectively; the last row demonstrates the feedforward torque adaptation to a 10 kg payload.}
  \label{fig:sim_exp1_1}
\end{figure}

\begin{figure}[t]
  \centering
  \includegraphics[width=1\linewidth]{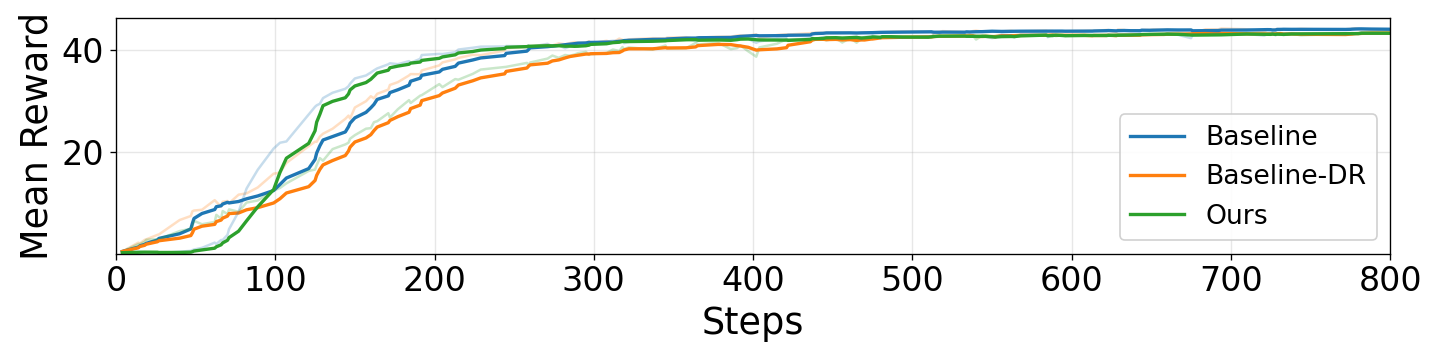} 
  \caption{Comparative results of learning efficiency.}
  \label{fig:sim_exp1_2}
\end{figure}

\begin{table}[t]
\centering
\caption{Sim-to-real Velocity Tracking Performance}
\label{tab:sim2real}
\scalebox{1.05}{
\begin{tabular}{lccc}
\toprule
Methods      & Isaacgym      & Pybullet & Real \\ \midrule
Baseline    &  $0.03 \pm 0.02$      & $0.14 \pm 0.03$ & $0.21 \pm 0.08$ \\
Baseline-DR &  $0.03 \pm 0.01$      & $0.12 \pm 0.03$ & $0.20 \pm 0.07$ \\
Ours        &  $0.04 \pm 0.01$       & $\textbf{0.07} \pm \textbf{0.05}$ & $\textbf{0.13} \pm \textbf{0.08}$ \\ \bottomrule
\end{tabular}}
\end{table}

Firstly, we demonstrate the isolation of reaction forces in action space and the dynamic adaptation ability benefited from the online MPC. As shown in the first two rows of Fig.~\ref{fig:sim_exp1_1}, there is a larger gap between action and joint position for the baseline in stance phase, indicating that the nominal RL method learns to resist the extra contact forces by increasing the joint position tracking error, leading to increased torque generated by the PD controller. In contract, the extra GRFs is provided by MPC in the form of feedforward torque in our method, and the joint action-position gap of our method is much smaller, revealing that the learned policy mainly rolls out the target position, a decoupled kinematics information. Note that with our method, the action is not perfectly aligned with the reference joint position which is purely kinematics information rolled out from the MPC module, because the MPC only finds a course dynamics trajectory, and the RL agent learns to improve upon it, represented by this gap, which validates our claim in Section~\ref{sec:introduction}. In addition, we validate our method's adaptation ability by putting a payload of 10 kg on the robot base to check how joint feedforward torque adapts to the unknown payload. As shown in the third row of Fig.~\ref{fig:sim_exp1_1}, MPC generates more feedforward torque when the payload is added to the robot, confirming the online adaptation ability benefiting from our decoupled framework. 

Secondly, the learning curves of the three methods are demonstrated in Fig.~\ref{fig:sim_exp1_2}, where our method achieves a more rapid convergence than the other two methods with a comparable mean reward, revealing the learning efficiency of our decoupled method achieved by shrinking the exploration space and shifting the focus to tracking gait motion. 

Finally, we validate the superiority of our method in sim-to-sim and sim-to-real transfer. As listed in Table~\ref{tab:sim2real}, all three methods perform almost equally in Isaac Gym, while when deployed in Pybullet which has a more accurate physics engine and in real world, our method outperforms both baseline and baseline-DR with the smallest root mean square of the velocity tracking error (RMSE). Note that the velocity in real world is estimated using a state estimator, which may not be very accurate, but it can roughly reflect the actual velocity. The close of the sim-to-real gap benefits from our insight that the correlation of reaction force control and gait motion control in the traditional RL control scheme may lead to a severer sim-to-real issue, and our method mitigates this problem successfully by decoupling force and position control.

\subsubsection{Horizontal Disturbances}

\begin{table}[t]
\centering
\caption{Comparison of Locomotion Performance under Horizontal Disturbances}
\label{tab:horizontal disturbance}
\begin{tabular}{l l c c}
\toprule
\multirow{2}{*}{Directions (x,y)} & \multirow{2}{*}{Methods} & \multicolumn{2}{c}{\textbf{Metrics}} \\
\cmidrule(lr){3-4}
& & Max. force ($N$) & RMSE ($m/s$)\\
\midrule
\multirow{3}{*}{(1, 0)} 
& Baseline      & 60           & $0.39 \pm 0.22$ \\
& Baseline-DR   & 60           & $0.45 \pm 0.25$ \\
& Ours          & \textbf{105} & $\textbf{0.26} \pm \textbf{0.10}$ \\
\midrule
\multirow{3}{*}{(-1, 0)} 
& Baseline      & 50           & $0.47 \pm 0.13$ \\
& Baseline-DR   & 40           & $0.46 \pm 0.13$ \\
& Ours          & \textbf{75}  & $\textbf{0.20} \pm \textbf{0.08}$ \\
\midrule
\multirow{3}{*}{(0, 1)} 
& Baseline      & 20           & $0.24 \pm 0.06$ \\
& Baseline-DR   & 25           & $0.26 \pm 0.06$ \\
& Ours          & \textbf{40}  & $\textbf{0.15} \pm \textbf{0.05}$ \\
\bottomrule
\end{tabular}
\end{table}

In this experiment, continued horizontal force is applied to the robot center of mass (CoM) from the front, side, and back of the robot, respectively. The disturbance force increases with a time interval of 5 seconds and a step of 5 \textit{N} until the robot falls, starting from 10 \textit{N}. Each of the three methods is run 10 times. We evaluate their performance using the maximum force that the robot can resist successfully for the whole time interval and the velocity tracking RMSE. The results are shown in Table~\ref{tab:horizontal disturbance}, from which we can conclude that our method achieves the best performance against horizontal disturbances applied from different directions. More specifically, baseline and baseline-DR perform worse in both velocity tracking and maximum forces that they can resist, because they can only adapt to the disturbances that they have experienced in the simulation during training. Moreover, baseline-DR can resist a bit larger lateral force than baseline as a heavier DR is implemented during the training of the former. Nevertheless, both baselines fail under the lateral force exceeding 25 \textit{N} that severely deviate from the simulation environment. In contrast, our method keeps the tracking errors small even under the large disturbances that cause the other two policies to fail, and this is achieved with a minimum DR in training, thanks to the decoupled framework that provides dynamically optimized feedforward GRFs.

\subsubsection{Locomotion on Complex Terrains}

\begin{table}[t]
\centering
\caption{Comparison of Locomotion Performance on Complex Terrains}
\label{tab:complex terrains}
\begin{tabular}{l c c}
\toprule
\multirow{2}{*}{Methods} & Uneven & Slope \\
\cmidrule(lr){2-3}
& Max. height range (m) & Max. slope (degree) \\
\midrule 
Baseline      & 0.05           & 5 \\
Baseline-DR   & 0.06           & 7 \\
Ours          & \textbf{0.07}  & \textbf{13} \\
\bottomrule
\end{tabular}
\end{table}

\begin{figure}[t]
  \centering
  \includegraphics[width=1\linewidth]{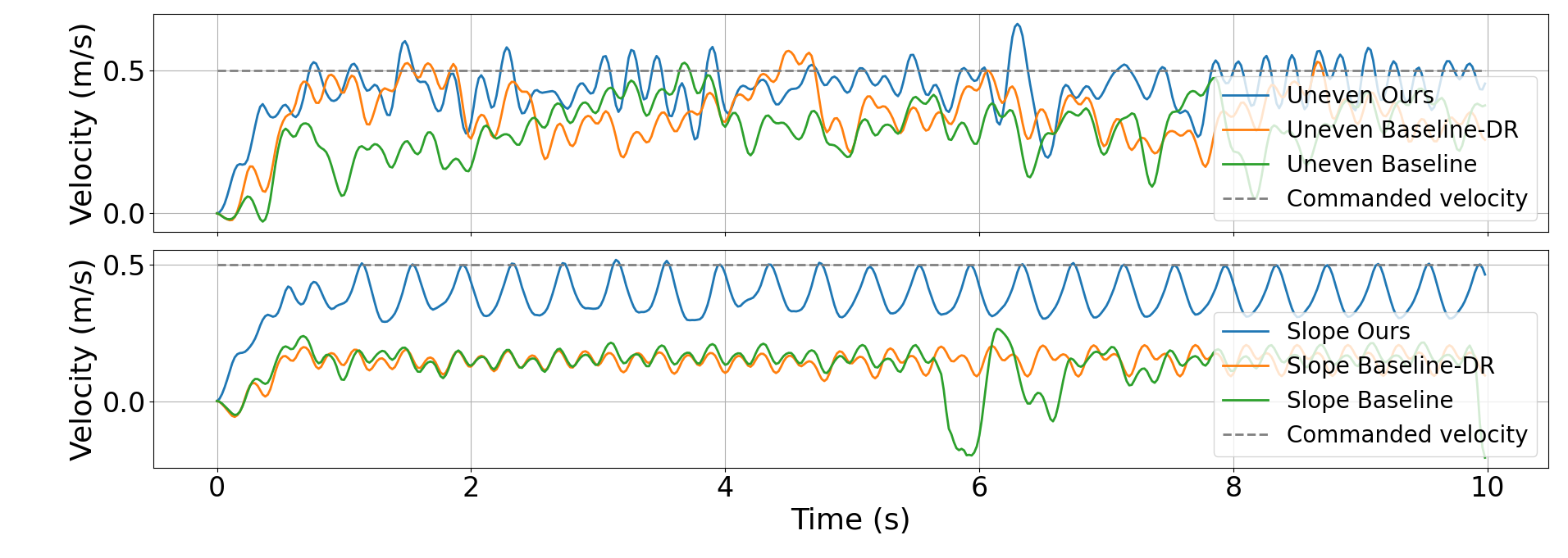} 
  \caption{Locomotion velocity on the uneven terrain with height range of 0.05 m and the inclined terrain with slope \(5^\circ \).}
  \label{fig:sim_exp3_1}
\end{figure}

In order to validate the generalization capability of our method across different terrains, we implement the three methods on an uneven terrain and a sloped planes as shown in Fig.~\ref{fig:show_photo}, respectively. We gradually increase the difficulty of the terrains, including the range of height variation for the uneven terrain and the angle (in degrees) of the sloped terrain. Table~\ref{tab:complex terrains} shows the maximum difficulty level that each method can successfully traverse for 10 s, and our method achieves the best performance among the three methods. Note that all three policies are trained on plane ground with the apex foot height of the swing phase set as 8 cm. Fig.~\ref{fig:sim_exp3_1} compares the velocities in base x-axis for different methods, and our method walks at the velocity most close to the commanded velocity on each terrain. This is because walking in complex terrains requires additional GRFs to overcome the impacts of environments. For example, when walking on the uneven terrain the robot's feet may get trapped by obstacles or get slipped by the steep edges, which reduces the components of GRFs on the heading direction, and walking on a sloped plane requires extra GRFs to overcome the rise of base altitude. These factors make the baseline methods walk slower because the simulation setups during training do not account for these environments, and their learned plane-ground dynamics degrades on the complex terrains. In contrast, our method utilizes an online MPC that finds optimal feedforward GRFs dynamically to the changing environment in real time, enhancing the adaptability to unforeseen environments.

\subsection{Hardware Experiments}
\subsubsection{Heavy Payloads}

\begin{figure}[t]
  \centering
  \includegraphics[width=0.95\linewidth]{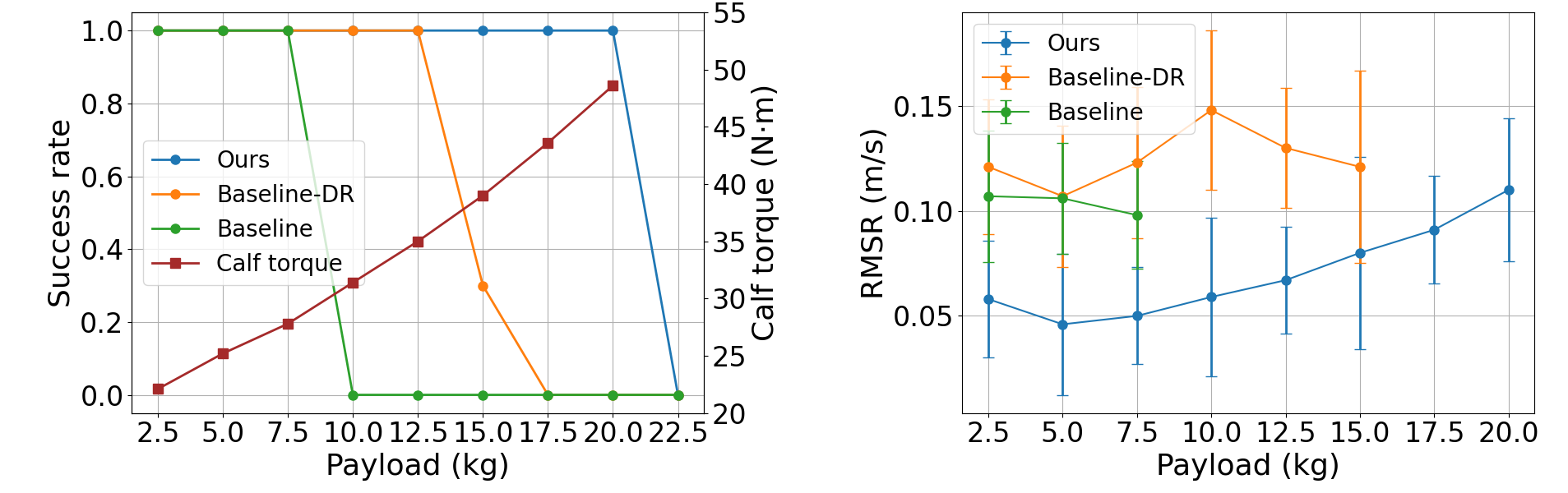}` 
  \caption{Success rate, mean calf feedforward torques, and tracking errors of commanded velocity.}
  \label{fig:real_exp1_1}
\end{figure}

\begin{figure}[t]
  \centering
  \includegraphics[width=0.95\linewidth]{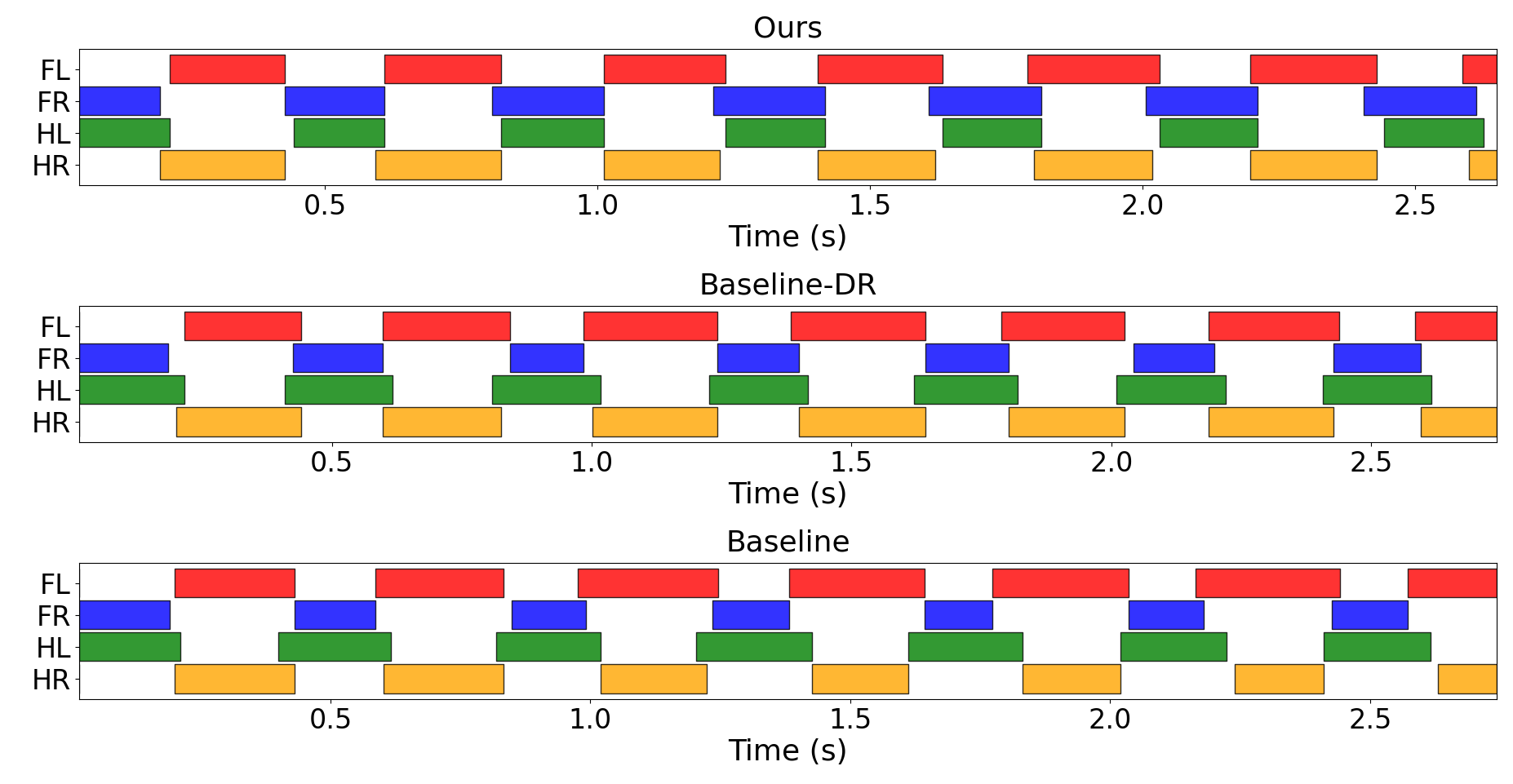}` 
  \caption{Gait sequence under the payload of 10 kg.}
  \label{fig:real_exp1_2}
\end{figure}

We evaluate the resistance capacity to disturbances of the three methods by adding payloads to the base of the robot that follows a velocity command of (0.3, 0, 0) m/s, and the weight of the payloads starts from 2.5~kg with an increment step of 2.5~kg until the robot falls within 10 s. We test 10 times for each method under each specific payload weight. We quantify the performance of different methods using success rate (SR) and RMSE of velocity tracking errors. As shown in Fig.~\ref{fig:real_exp1_1}, our method achieves the best performance in both payload and velocity tracking errors, aligning with the results of previous simulation experiments. The superiority of our method to adapt to unknown payloads can also be demonstrated in Fig.~\ref{fig:real_exp1_2}, where gait sequences using different methods with a payload of 10 kg are presented. Note that the nominal gait is trot with a gait cycle of 0.4 s and a duty cycle of 0.5. When carrying a 10 kg payload, the gaits of baseline and baseline-DR deviate from the nominal gait, representing by a longer stance phase for the front left (FL) leg and a shorter stance phase for the front right (FR) leg, while the gait with our method is closer to the nominal gait. This is because our decoupled method is able to adapt to the payloads dynamically, as shown in the variation of the mean calf feedforward torque of the four legs in Fig.~\ref{fig:real_exp1_1}, where the feedforward torque grows approximately linearly with the payloads, and this enables to correct the robot states to its desired states.

\begin{table}[t]
\centering
\caption{Comparison of Locomotion Performance with Biased Payloads}
\label{tab:biased payloads}
\begin{tabular}{l l c c c}
\toprule
\multirow{2}{*}{Methods} & \multirow{2}{*}{Payloads} & \multicolumn{3}{c}{\textbf{Metrics}} \\
\cmidrule(lr){3-5}
& & SR & RMSE ($v_\text{x}$) ($m/s$) & RMSE ($v_\text{y}$) ($m/s$)\\
\midrule
\multirow{3}{*}{Baseline} 
& 2.5 $kg$      & 0             & /           & / \\
& 5.0 $kg$      & 0             & /           & / \\
& 7.5 $kg$      & 0             & /           & / \\
\midrule
\multirow{3}{*}{Baseline-DR} 
& 2.5 $kg$      & 1             & $0.11 \pm 0.06$   & $0.16 \pm 0.07$ \\
& 5.0 $kg$      & 0             & /            & / \\
& 7.5 $kg$      & 0             & /            & / \\
\midrule
\multirow{3}{*}{Ours} 
& 2.5 $kg$      & 1             & $0.03 \pm 0.02$   & $0.02 \pm 0.01$ \\
& 5.0 $kg$      & 1             & $0.04 \pm 0.03$   & $0.02 \pm 0.01$ \\
& 7.5 $kg$      & 1             & $0.05 \pm 0.03$   & $0.03 \pm 0.03$ \\
\bottomrule
\end{tabular}
\end{table}

\subsubsection{Biased Payloads}

\begin{figure}[t]
  \begin{subfigure}[t]{0.325\columnwidth}
    \includegraphics[width=\linewidth]{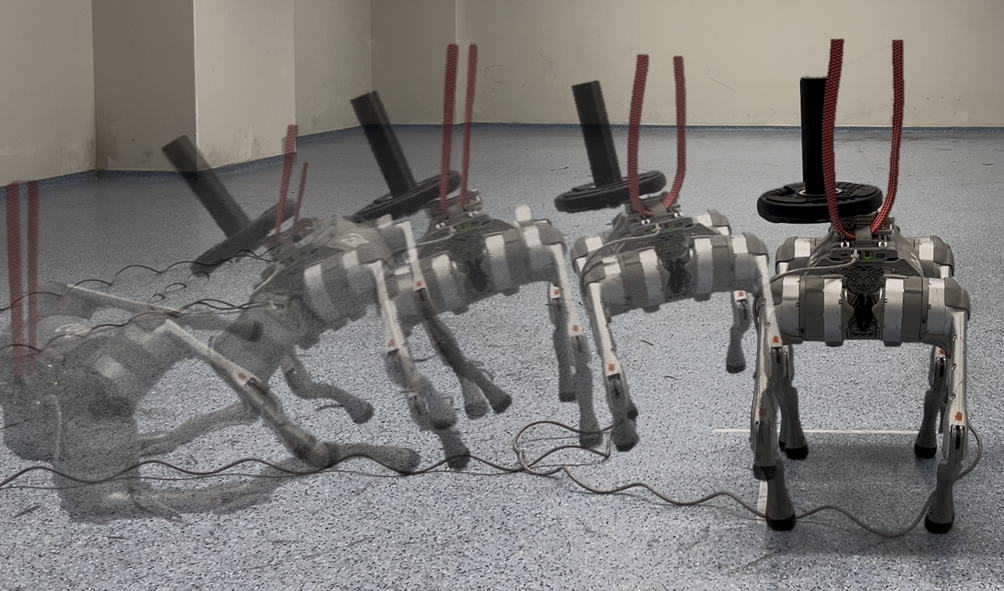}
    \subcaption{Baseline}
    \label{fig:real_exp2_3_baseline}
  \end{subfigure}
  \begin{subfigure}[t]{0.325\columnwidth}
    \includegraphics[width=\linewidth]{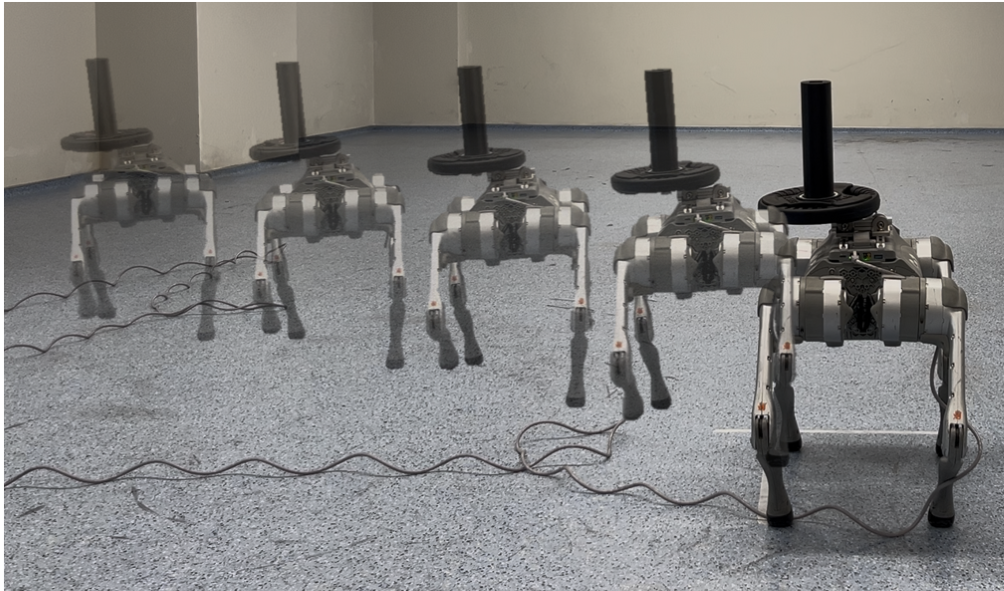}
    \subcaption{Baseline-DR}
    \label{fig:real_exp2_3_DR}
  \end{subfigure}
  \begin{subfigure}[t]{0.325\columnwidth}
    \includegraphics[width=\linewidth]{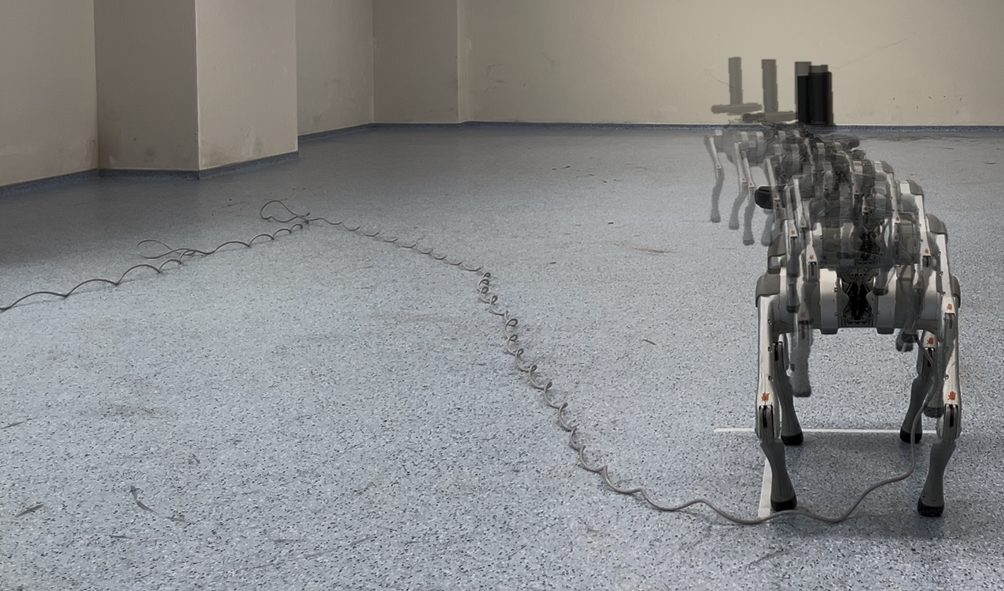}
    \subcaption{Ours}
    \label{fig:real_exp2_3_ours}
  \end{subfigure}
  \caption{Experiment snapshots with biased payloads of 2.5 kg}
  \label{fig:real_exp2_3}
\end{figure}

\begin{figure}[h]
  \centering
  \includegraphics[width=0.95\linewidth]{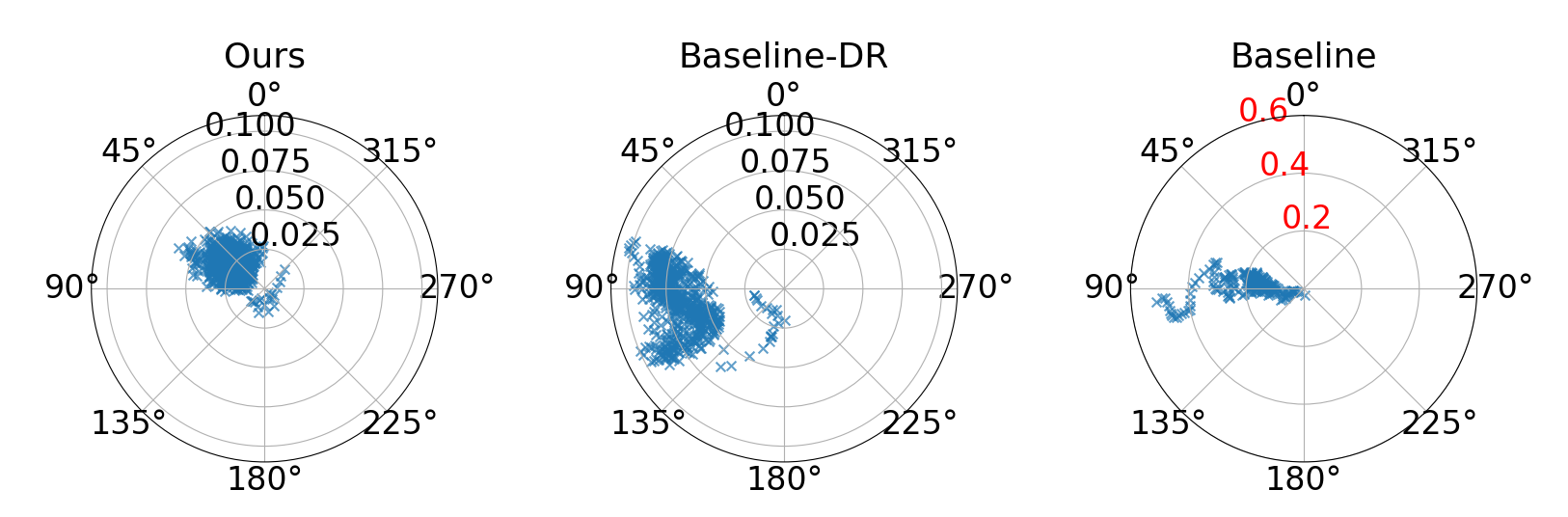}` 
  \caption{Gravity projection of the comparative methods.}
  \label{fig:real_exp2_1}
\end{figure}

In this experiment, we shift the payloads 7 cm along the base y-axis from the base origin. This will apply an extra roll moment on the robot base, which increases the challenge of the disturbances. Apart from the change in the position of payloads, experiment setups keep the same as in the heavy-load experiment. As shown in Table~\ref{tab:biased payloads}, our method outperforms both baseline and baseline-DR not only in the maximum weight of payloads that can be carried successfully, but also in the accurate tracking of the commanded velocity. More specifically, experiment snapshots carrying a 2.5 kg payload are demonstrated in Fig~\ref{fig:real_exp2_3}, and it shows that without the online adaptive MPC, the robot cannot resist the roll moment and gradually deviates to the left, while our method utilizes online MPC to resist the moment and walks more straight. The effectiveness of our method is also represented by a more normal base orientation, which is validated by more centralized gravity projections, as illustrated in Fig.~\ref{fig:real_exp2_1}. 

\section{Conclusion}

In this work, a decoupled control framework is proposed, which assigns stance-leg control and swing-leg control to MPC and RL, respectively. This decoupling shrinks the exploration space for the RL agent by isolating the correlation of the two control spaces, thus, achieves efficient learning and closes the sim-to-real gap. Moreover, dynamic adaptation in unfamiliar environments is achieved by stance-leg control module that optimizes GRFs using an online MPC during deployment. Various simulation and real-world experiments validate its performance against baselines. Future work includes replacing the swing trajectory generator using Raibert's heuristic with a more flexible method to enable the traversability on challenging terrains.


\bibliographystyle{IEEEtran}
\bibliography{ref}

\addtolength{\textheight}{-12cm}   

\end{document}